\documentclass[10pt,twocolumn,letterpaper]{article}

\usepackage{wacv}
\usepackage{times}
\usepackage{epsfig}
\usepackage{graphicx}
\usepackage{amsmath}
\usepackage{amssymb}
  
\usepackage{times}
\usepackage{soul}
\usepackage{url}
\usepackage[utf8]{inputenc}
\usepackage[small]{caption}
\usepackage{graphicx}   
\usepackage{amsmath}
\usepackage{booktabs}
\usepackage{algorithm}
\usepackage{algorithmic}
\usepackage{microtype}
\usepackage{subfigure}
\usepackage{times}
\usepackage{epsfig}
\usepackage{amssymb}
\usepackage{multirow}
\usepackage{svg}
\usepackage{graphics}
\usepackage{tabularx}

\wacvfinalcopy 

\def\wacvPaperID{****} 

\ifwacvfinal\fi
\begin{document}

\title{Robust Explanations for Visual Question Answering}

  \author{ {Badri N. Patro}  \quad {Shivansh Patel}  \quad  {Vinay P. Namboodiri} \\
  Indian Institute of Technology, Kanpur \\
  {\tt \{badri,shivp,vinaypn\}@iitk.ac.in} \\}

\maketitle
\ifwacvfinal\thispagestyle{empty}\fi

\begin{abstract}
In this paper, we propose a method to obtain robust explanations for visual question answering(VQA) that correlate well with the answers. Our model explains the answers obtained through a VQA model by providing visual and textual explanations. The main challenges that we address are i) Answers and textual explanations obtained by current methods are not well correlated and ii) Current methods for visual explanation do not focus on the right location for explaining the answer. We address both these challenges by using a collaborative correlated module which ensures that even if we do not train for noise based attacks, the enhanced correlation ensures that the right explanation and answer can be generated. We further show that this also aids in improving the generated visual and textual explanations. The use of the correlated module can be thought of as a robust method to verify if the answer and explanations are coherent. We evaluate this model using VQA-X dataset. We observe that the proposed method yields better textual and visual justification that supports the decision. We showcase the robustness of the model against a noise-based perturbation attack using corresponding visual and textual explanations. A detailed empirical analysis is shown. Here we provide source code link for our model \url{https://github.com/DelTA-Lab-IITK/CCM-WACV}.
\end{abstract}

\section{Introduction}
\begin{figure}[h]
	\centering
	\includegraphics[width=0.46\textwidth]{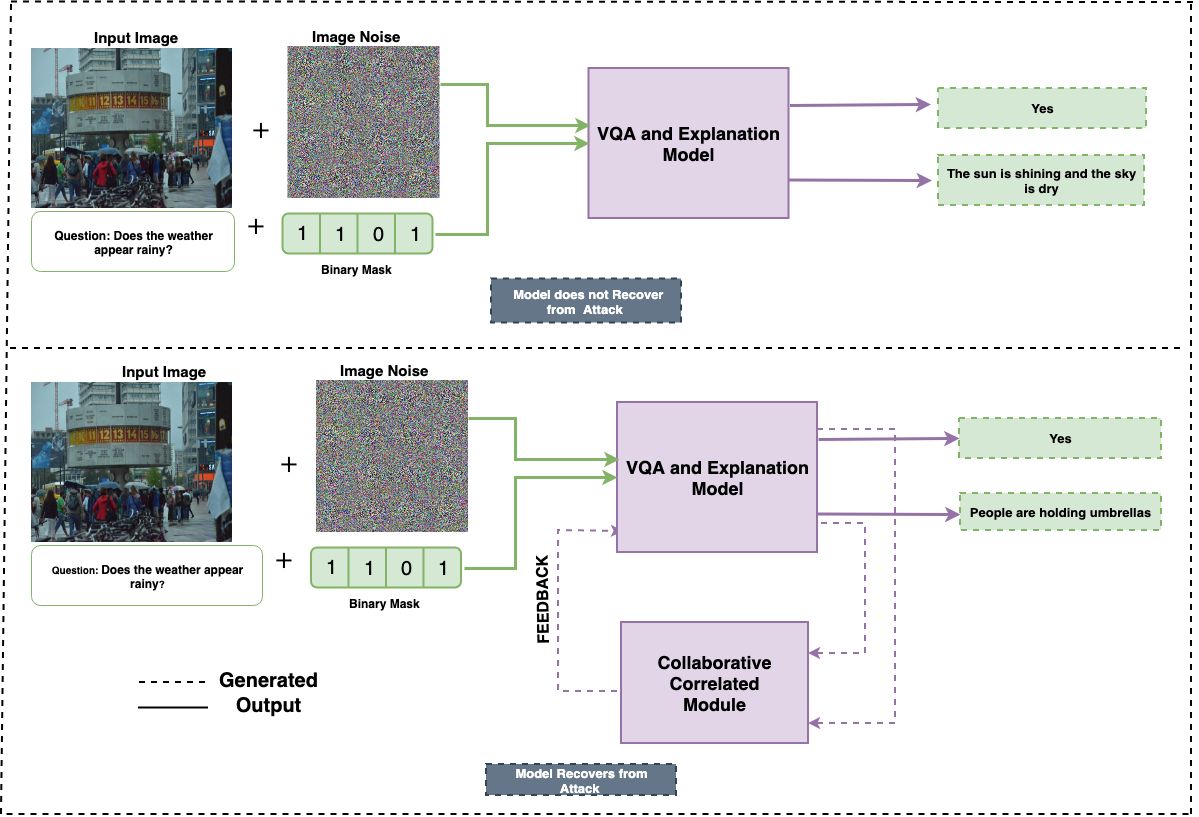}
	\vspace{-0.5em}
	\caption{Illustration of proposed method: In case coherence of explanation and answer generation of VQA network is not enforced, a noise based perturbation will result in diverse answer and explanation being generated. This is shown in first row. In second row we illustrate that the proposed method ensures coherence and therefore is able to be robust to noise based perturbation.} 
	\label{fig:attack}
	\vspace{-1em}
\end{figure}
In this paper, we solve for obtaining robust explanations for visual question answering. Visual question answering is a semantic task that aims to answer questions based on an image. The practical implication for this task is that of an agent answering questions asked by, for instance, a visually impaired person that wants to know answers. An important aspect related to this model is the ability to reason whether the model is able to really understand and provide explanations for its answer. This aspect was investigated, for instance, by a recent method by  \cite{huk_CVPR2018multimodal}, where the authors proposed a method to generate textual explanations and also provide localizations that contribute to the explanation for their answer. This was obtained by their approach which generates explanations based on the answer attention. However, one drawback we observe for such an approach is that the explanation need not be correct. For instance, using a noise based perturbation on the image, we can have instances of answer and explanation being different. We solve this by jointly generating the answer and explanation. We further improve over this method by proposing a novel method that enhances the correlation by verifying that the answer and explanation do agree with each other. This is obtained through a collaborative correlated network. This concept is illustrated in figure~\ref{fig:attack} where we show that current methods can generate answers and explanations. However, these may diverge for an image corrupted using noise based perturbation. Our proposed method aids in generating robust explanations that are tolerant of such perturbation though our methods are not trained using such noise at the time of training.
   
\begin{figure}[ht]
	\centering
	\includegraphics[width=0.47\textwidth]{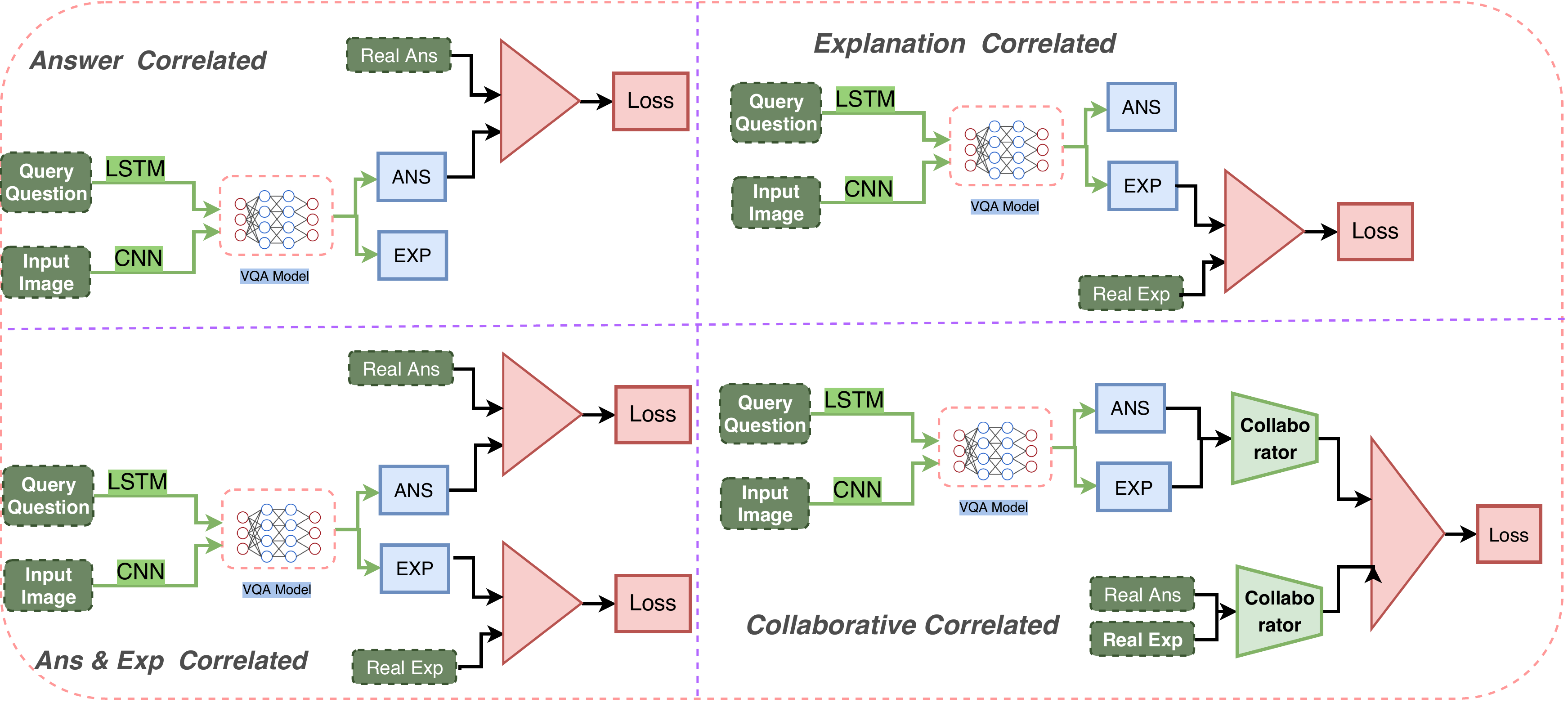}
	\caption{ This figure shows variations of our methods. Answer Correlated only corrects the altered answers. Explanation Correlated only corrects the altered explanations. Answer and Explanation Correlated corrects answer and explanation separately. Collaborative Correlated Module jointly corrects both altered answer and explanation.}
	\label{fig:over}
\end{figure}

We investigate various other ablations of the system, such as verifying whether the answer is correct or the explanation is correct or separately verifying that the answer and explanations are correct. These variations are illustrated in figure~\ref{fig:over}. We observe that jointly verifying that the answer and explanations are correct and agree with each other is better than all the other variations. This is termed as a collaborative correlation module as the answer and explanation collaboratively are checked. This module not only aids in generating better textual explanations but also helps to generate better localization for the visual explanation. It also ensures the coherence of answer and explanation generation.

To summarize, through this paper we provide the following contributions:
\begin{itemize}
    \item We investigate the tolerance of VQA systems against noise based perturbation and propose a method that is robust to such perturbation. 
    \item We propose joint explanation and answer generation systems that are improved using a novel collaborative correlated network that ensures that the answer and explanations are coherent and correct. 
    \item We illustrate the role of the correlated networks by providing corresponding attention maps that correlate well with human annotated attention maps.
    \item A detailed empirical analysis of variants of our model provides quantitative evidence of improvements over the current state of the art methods in terms of improved answers being tolerant to adversarial attacks and generating coherent textual and visual explanations.
\end{itemize}    

\section{Related Work}\label{sec:lit_surv}
Extensive work has been done in the Vision and Language domain for solving image captioning~\cite{Barnard_JMLR2003,Farhadi_ECCV2010,Kulkarni_CVPR2011,Socher_TACL2014,Vinyals_CVPR2015,Karpathy_CVPR2015,Xu_ICML2015,Fang_CVPR2015,Chen_CVPR2015,Johnson_CVPR2016,Yan_ECCV2016}, Visual Question Answering (VQA) ~\cite{Malinowski_NIPS2014, Lin_ECCV2014,VQA, Ren_NIPS2015, Ma_AAAI2016, Noh_CVPR2016}, Visual Question Generation (VQG) ~\cite{Mostafazadeh_ACL2016, jain_CVPR2017,Patro_EMNLP2018MDN} and Visual Dialog ~\cite{Das_ICCV2017,Aytar_TPAMI2017,Velivckovic_SSCI2016,Vijayakumar_2016diverse,Yu_ICCV2015}. Malinowski{\textit et al.} ~\cite{Malinowski_NIPS2014} has proposed Visual question answering task, which answer natural language question based on the image. ~\cite{VQA,Goyal_CVPR2017} generalize this task with a large bias free dataset. Joint embedding approach was proposed by~\cite{VQA,Lin_ECCV2014,VQA,Ren_NIPS2015,Ma_AAAI2016,Noh_CVPR2016} where they combine image features with question features to predict answers. Attention-based approach comprises image-based attention, question-based attention and both image and question based attention. Recent work from~\cite{Zhu_CVPR2016,Fukui_arXiv2016,Gao_NIPS2015,Xu_ECCV2016,Lu_NIPS2016,Shih_CVPR2016,Li_NIPS2016,Patro_2018_CVPR,Patro2019_ICCV2019UCAM} considers region-based image attention.

\textbf{Explanation}:
Early textual explanation models spanned a variety of applications such as medicine~\cite{shortliffe_MB1975model}, feedback for teaching~\cite{lane_USC2005explainable} and were generally template based. Some methods find discriminative visual patches~\cite{doersch_ACM2012makes},  \cite{berg_ICCV2013you} whereas others  aim to understand intermediate features which are important for the end decisions~\cite{zeiler_ECCV2014visualizing}, ~\cite{escorcia_CVPR2015relationship}, \cite{zhou_arxiv2014object}. Recently, the author ~\cite{Patro2019_ICCV2019UCAM} has proposed a new paradigm of providing visual explanation using uncertainty based class activation map. Furthermore, a variety of works proposed methods to visually explain decisions. The method that is closest to our work is the recent work by ~\cite{huk_CVPR2018multimodal}. Their work aims at generating explanations and also providing visual justification for their answer.  None of the methods checks robustness. To the best of our knowledge, we are raising a new novel issue related to the robustness of the prediction and explanation that has not been previously considered in the literature.  Our work is inspired by this effort and aims to provide a more robust and coherent explanation generation, which is verified experimentally.

\textbf{Adversarial Methods}: Generative adversarial networks (GAN)~\cite{goodfellow_NIPS2014} is an adversarial methods that consists of a generative model, G, which captures the data distribution, and a discriminative model, D, which estimates the probability of a sample as to whether it came from the training data or not. GANs are widely used to explain data distribution and various other tasks in vision domain ~\cite{mirza_arxiv2014conditional,radford2015unsupervised}. Also, there has been a lot of work on GANs in the field of natural language processing ~\cite{mirza_arxiv2014conditional,zhang_arxiv2017adversarial,Hu_ICML2017toward,Yu_AAAI2017seqgan,Guo_arxiv2017long,liang_CORR2017recurrent}. Reed {\it et al.} \cite{Reed_CVPR2016} have proposed a model which combines vision with language to generate image from text. Li {\it et al.}~\cite{li_arxiv2017adversarial} has proposed an advanced method to generate sequence of conversion about an image. Patro {\it et al.}~\cite{Patro2019ExplanationVA} have proposed an adversarial method to improve explanation and attention using surrogate supervision method. \\
In this work, we propose a collaborative correlated module to generate both answer and textual  explanation of that answer which will be tightly correlated with each other. We show that this module ensures that even if we do not train for noise based attacks, the enhanced correlation can ensure that the right explanation and answer can be generated.

   \begin{figure*}[ht]
	\centering
	\vspace{-0.5cm}
	\includegraphics[width=1.0\textwidth]{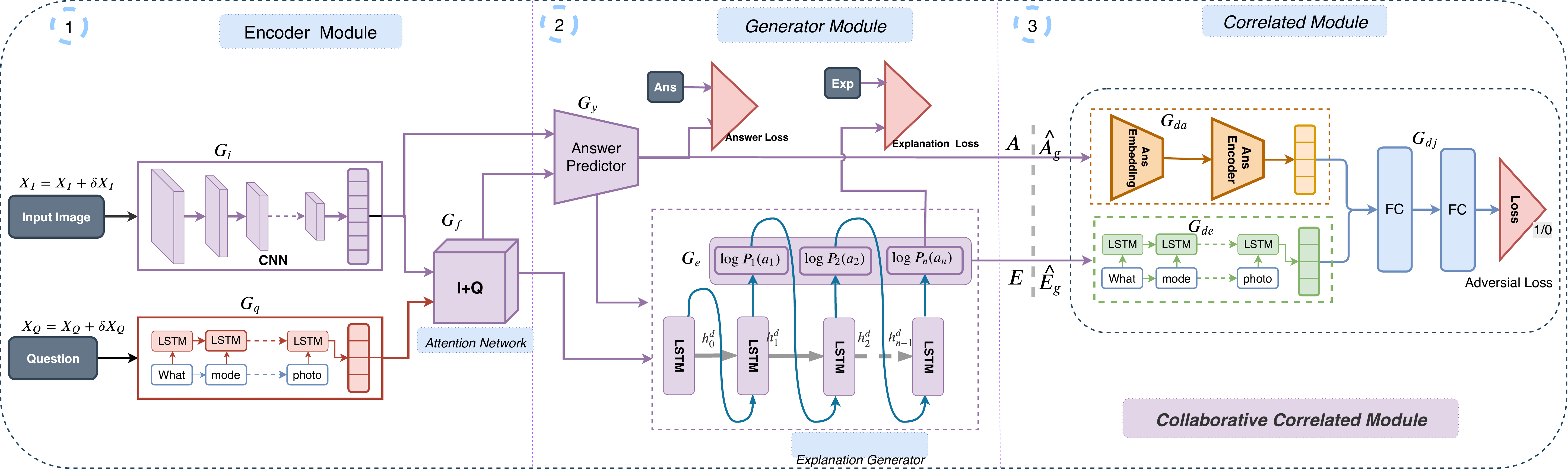}
	
	\caption{Illustration of Collaborative Correlated Model(CCM). The model receives input image feature and its question feature using CNN and LSTM respectively. then the model predict answer and generate explanation for the predicted answer. During training we ensure that the model collaborates with both answer and explanation features and learns jointly using adversarial fashion.}\label{fig:main}
\end{figure*}
\section{Method}
Interpretability of models through explanation does not consider the robustness and consistency of the provided explanations. While providing interpretability for understanding models is important, ensuring that the same are robust and consistent is also crucially important. We are specifically investigating this problem and provide a simple framework to solve for the same.
The main focus of our approach is to ensure correctness and  correlation in visual question answering(VQA). We propose a correlated network that learns joint embedding by collaborating with textual explanation and answer embedding. The key difference between our architecture and other existing VQA architectures is in the use of a mutually collaborating module for explanation and answering blocks in a joint adversarial mechanism. This is illustrated in figure~\ref{fig:main}. The other aspects of VQA  and explanation are retained as it is. In particular, we adopt a classification based approach for solving VQA where an image embedding is combined with the question embedding to solve for the answer. This is done using a softmax function in a multiple choice setting:

\[\hat{A}=\underset{A \in \Omega}{argmax}P(A|I,Q : \theta)\] where $\Omega$ is a set of all possible answers, I and Q are image and question respectively and $\theta$ represents the parameters in the network. Also we adopt generation based approaches for 
explanation generation \[\hat{E}=\underset{E \in \Omega_e}{argmax}P(e_t|(I,Q), A, e_0, \dots e_{t-1} : \theta_e)\] where $\Omega_e$ is the explanation vocabulary.\\
We provide four variants of our model, as shown in figure \ref{fig:over}. Correlated Answer Module (CAM) only corrects altered answers, Correlated Explanation Module (CEM) only corrects altered explanations, Answer and Explanation Correlated Module (AECM) corrects answers and explanations separately and Collaborative Correlated Module (CCM) corrects both answers and explanations jointly.\\
CCM comprises of three modules: Encoder Module,  Generator Module, and Correlated Module. The Encoder Module encodes images and questions using CNN and LSTM. We combine these two using an attention mechanism to obtain an attention feature. Using the attention feature, the generator module predicts answers and explanations that justify the predicted answer. The Correlated module defends the model against perturbations in the input, which facilitates the model to predict correct answers even under perturbation though we do not perturb the input at the time of training. Details of each module are as follows:


\subsection{Encoder Module}
\vspace{-0.3cm}
Given an input image $X_i$, we obtain an embedding $g_i \in \mathcal{R}^{W \times H \times C}$ using CNN which is parameterized by a function $G_i(X_i,\theta_i)$, where $\theta_i$ represents the weights for the image embedding module. Similarly, for the query question $X_Q$, we obtain question feature embedding $g_q$  after passing it through an LSTM, which is parameterized using a function $G_q(X_q,\theta_q)$, where $\theta_q$ represents the weights for the question embedding module. The image embedding $g_i$ and question embedding $g_q$ are used in an attention network, which is  parameterized using a function $G_f(g_i,g_q,\theta_f)$  where $\theta_f$ are the parameters of the attention network that combines the image and question embeddings with a weighted softmax function and produces an output attention weighted vector $g_f$. The corresponding attention expressions are as follows:

\begin{equation}
    \begin{split}
        & f_{j}=\tanh(\theta_{i}g_i \odot \theta_{q} g_q)\\
        & f_{s}=||(\textit{signed\_sqrt}(f_{j}))||_2\\
        &\alpha = \text{softmax}(\theta_{a1}\sigma(\theta_{a}f_{s} +b_{a}))\\
        & g_{f} = (\alpha*g_i) \odot f_q\\
    \end{split}
\end{equation}
where $\odot$ represents element-wise multiplication and $\sigma(.)$ represents the sigmoid operator.


\subsection{Generator Module}
\vspace{-0.3cm}
Our generator module aims to predict an answer and generate an explanation. Both the answer prediction and textual explanation rely on the attention feature $g_f$. Answer generation is through a classifier network $g_y$ using a fully connected network. 
Our generator module aims to predict answer and generate explanation. First, it helps to obtain the probability of the predicted answer class with the help of a softmax classifier. The answer prediction relies on the attention feature $g_f$. $g_f$ is projected into $V_a$ dimensional answer space using a fully connected layer. Answer classifier network is defined as follows: $g_y=h(\theta_{y_1},h(\theta_{y_2},g_f))$
\begin{equation}
      g_y=h(\theta_{y_1},h(\theta_{y_2},g_f))
\end{equation}
where $h$ is the ReLU activation function, $\theta_{y_1} \in \mathcal{R}^{V_a \times l1} $ and $ \theta_{y_2} \in \mathcal{R}^{l1 \times l_{f}}$.
At training time, we minimize the cross entropy loss between the predicted and the ground truth answers.
\begin{equation}
         L_y(\theta_{f},\theta_y)=L_y(G_y(G_f(g_i,g_q));y)
\end{equation}
For generating textual explanation, we condition attention feature $g_f$ on answer embedding $g_y$. This is generated using an LSTM based sequence generator for generating textual explanation $g_e$. At training time, we generate meaningful explanations by minimizing the cross entropy loss between generated explanation and ground truth explanation. 

\begin{equation}
 L_e(\theta_{f},\theta_y,\theta_e)=L(G_e(G_y,G_f);e)\\
\end{equation}

\subsection{Correlated Module}\label{mod:def}
We introduce an Adversarial Correlated Network. Using the adversarial mechanism, we develop a Collaborative Correlated Module which simultaneously checks whether the predicted answer and the corresponding explanation are correct or not. Each time the model predicts a wrong answer, the correlated module tries to correct it by comparing it with the real answer and the corresponding real explanation. Our correlated module includes Correlated Answer  Module, Correlated Explanation Module and Collaborative Correlated Module. Each of the correlated modules are explained as follows:

\textbf{ Correlated Answer Module}:
 We use a $V_a$ dimensional one-hot vector representation for every answer word and transform it into a real valued word representation $f_{da}$ by matrix $\theta_{aw} \in \mathcal{R}^{l_{aw} \times v_a}$. We pass the obtained $l_{aw}$ dimensional word embedding through a fully connected layer to obtain $l_a$ dimensional answer encoding $g_{da}$. $g_{da}$ is represented as follows: $g_{da}=h(\theta_{da},h(\theta_{aw},A))$, where $h$ is a nonlinear function ReLU.
\begin{algorithm}[htb]
  \caption{Training CCM}
  \label{alg:example}
\begin{algorithmic}[1]
  \STATE {\bfseries Input:}  Image $X_I$, Question $X_Q$
  \STATE {\bfseries Output:}  Answer $y$, Explanation $e$
  \REPEAT
    \STATE  \text{Answer Generator  $ G_y(X_I, X_Q)\gets g_y$}
    \STATE  \text{Explanation Generator  $ G_e(X_I, X_Q)\gets g_e$}
    \STATE  \textit{Ans cross entropy $L_{ans} \gets$ loss$(\hat{y},y)$}
    \STATE  \textit{Exp cross entropy $L_{exp} \gets$ loss$(\hat{e},e)$}
    
    \REPEAT 
          \STATE \textit{Sample mini batch of fake Ans and Exp:\\
                         \hspace{2cm} $y_f^1 \dots y_f^m$  and $e_f^1 \dots e_f^m$}
          \STATE \textit{Sample mini batch of real Ans and Exp:\\
                       \hspace{2cm} $y_r^1 \dots y_r^m$ and  $e_r^1 \dots e_r^m$ }
            \STATE  \text{Discriminator: $ D_j(D_y(y_r^i),D_e(e_r^i))\gets D(Y,E)$}            
           \STATE \textit{Update the discriminator  using stochastic gradient ascent\\ $ \nabla_{\theta_d}\frac{1}{m}\sum_{i=1}^{m} [ \log D(Y,E) + \log(1-D(G_y,G_e))]$ }
           \UNTIL{$k=1:K$}
        \STATE \textit{Sample mini batch of Real Ans and Exp:\\
                        $y_f^1 \dots y_f^m$ and  $e_f^1 \dots e_f^m$ }
         \STATE \textit{Update the Generator  by descending its stochastic gradient: $ \nabla_{\theta_g}\frac{1}{m}\sum_{i=1}^{m} \log(1-D(G_y,G_e))$ }
    \UNTIL{Number of iterations}
\end{algorithmic}
\end{algorithm}

 \begin{table*}[ht!]
   \centering
   \begin{tabular}{|l|c|c|c|c|c|c|c|c|c|c|c|}
     \hline
     \textbf{Model}&\textbf{BLEU-1}&\textbf{BLEU-2}&\textbf{BLEU-3}&\textbf{BLEU-4}& \textbf{METEOR}& \textbf{ROUGE-L}& \textbf{CIDERr}& \textbf{SPICE}\\\hline
     Baseline & 54.7 & 38.1 & 26.8 & 19.1 & 18.0 & 42.9 & 66.1 & 14.0\\
     CAM & 55.0 & 38.3 & 26.8 & 19.0 & 18.3 & 43.1 & 69.2 & 15.2\\
     CEM & 54.5 & 38.3 & 27.2 & 19.6 & 18.4 & 43.1 & 68.2 & 15.1\\
     AECM & 55.5 & 38.9 & 27.4 & 19.3 & 18.3 & 43.4 & 67.9 & 14.8\\
     CCM & \textbf{56.7} & \textbf{40.8} & \textbf{29.2} & \textbf{21.1} & \textbf{19.7} & \textbf{44.9} & \textbf{73.9} & \textbf{16.2}\\\hline
   \end{tabular}
    \vspace{-1em}
   \caption{\label{abalation} Ablation Analysis of Our Model. We achieve improvements in all the metrics.}
 \end{table*}

\begin{figure*}[htb]
     \small
     \centering
     \begin{tabular}[b]{ c  c }
     (a) Image  & (b) Question \\ 
     \includegraphics[width=0.5\textwidth]{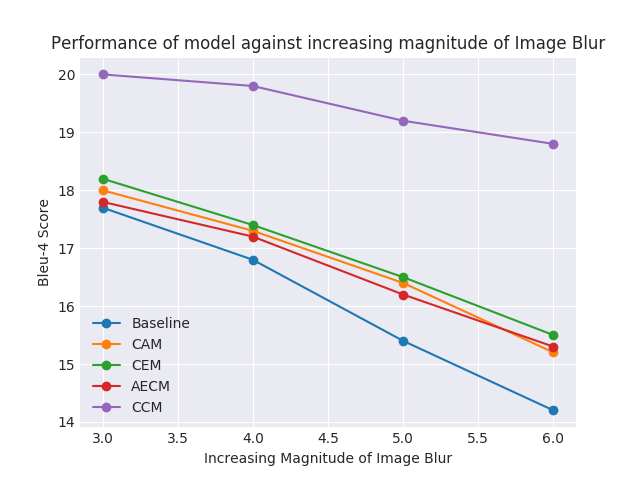}
     & \includegraphics[width=0.5\textwidth]{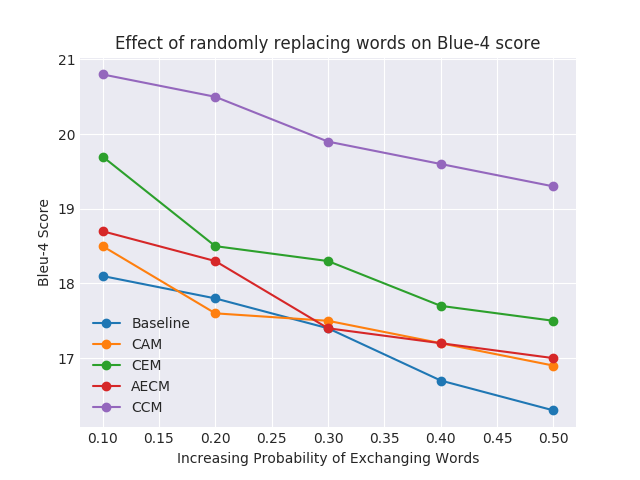}
       \end{tabular}
      \caption{(a)\textbf{Blurring Images}: BLUE-4 score of our variants of model vs increasing blur in validation images. Slope with which CCM's score decreases is less as compared to other models and hence it is robust to blur in images. (b)\textbf{Replacing Words}: BLUE-4 score of our variants of models vs increasing random exchange of question words with question vocabulary words. Slope with which CCM's score decreases is less as compared to other models and hence it is robust to replaced question words.	}
      \label{fig:replacing_words_image}
      \vspace{-1em}
 \end{figure*}
\textbf{Correlated Explanation Module :}
For explanation, we obtain representation $g_{de}$  using an LSTM. The hidden state representation of the last word of the LSTM network provides a semantic representation of the whole sentence conditioned on all the previously generated words $e_0, e_1....e_{t}$. The model can be  represented as $g_{de}= LSTM(E)$ 

\textbf{Collaborative Correlated Module:}\label{CCM}
We design a collaborative network by concatenating the answer and explanation embeddings to obtain a joint embedding. The collaborative module is trained in an adversarial setting to correct the misclassified answers. Given answer embedding $g_{da}$ and explanation embedding $g_{de}$, we obtain a joint feature $g_{dj}$ by concatenating both of them and passing it through a fully connected layer to obtain final feature embedding which is as follows: 
\[ g_{dj}=\theta_{dj}\tanh(\theta_{da}g_{da}; \theta_{de}g_{de})\]
where ; indicates the concatenation of two modules. The complete scheme is shown in figure~\ref{fig:main}. We use these modules to make variants of our model, as shown in figure \ref{fig:over}. 
We train the discriminator in an adversarial manner between the generated and ground truth embedding. The adversarial cost function is given by: 
\begin{equation*}
\begin{split}
& \min_{G} \max_{D} L_{c}(G,D) = E_{{y} \sim Y,{e} \sim E} [\log D(Y,E)] + \\
                    & E_{g_{y} \sim G_y, g_{e} \sim G_e} [\log (1- D(G_y,G_e))]
\end{split}
\end{equation*}
The final cost function for CCM can be formulated as follows:
\[L = L_{y} + L_{e} - \eta L_{c}\]
where $L_{y}$ is the loss of answer generator module, $L_{e}$ is the loss of explanation generator module, $L_{c}$ is the loss of collaborative correlated module and $\eta$ is a hyper-parameter.
We trained our model by optimizing this cost function with model parameters $(\hat{\theta}_f, \hat{\theta_{e}}, \hat{\theta}_y, \hat{\theta}_d)$ to deliver a saddle point function as follows:
	\begin{equation}
    \begin{split}
        & (\hat{\theta}_{f},\hat{\theta}_{e},\hat{\theta}_y)= \arg\max_{\theta_f,\theta_e,\theta_y}(C(\theta_f,\theta_e, \theta_y,\hat{\theta}_d))\\
        & (\hat{\theta}_{d})= \arg\min_{\theta_d}(C(\hat{\theta}_f,\hat{\theta}_e,\hat{\theta}_y,\theta_d))\\
    \end{split}
\end{equation}




\section{Experiments}
We evaluate our proposed CCM method using quantitative and qualitative analysis. The quantitative evaluation is conducted using standard metrics like BLEU ~\cite{Papineni_ACL2002}, METEOR~\cite{Banerjee_ACL2005}, ROUGE~\cite{Lin_ACL2004} and CIDEr~\cite{Vedantam_CVPR2015}. We evaluate our attention maps using rank correlation ~\cite{Patro_2018_CVPR}. We further consider the statistical significance for the many ablations as well as the state-of-the-art models. In qualitative analysis, we show the word statistics of generated explanation with a Sunburst plot in figure~\ref{fig:sunburst}. 
We provide gradual improvement in visualization of attention maps for a few images as we move from our base model to CCM model. We perform ablation analysis based on various samples of noise. 

 \begin{table}[ht]
            \centering
 	  \begin{tabular}{|l|c|}
     \hline
     \textbf{Model}& \textbf{RC}\\\hline
     Random Point \cite{huk_CVPR2018multimodal} & +0.0017\\
     Uniform \cite{huk_CVPR2018multimodal} & +0.0003\\
     Answering \cite{huk_CVPR2018multimodal} & +0.2211\\
     ME \cite{huk_CVPR2018multimodal}  & +0.3423\\\hline
     Baseline (ME)  & +0.3425\\
     CAM (ours) & +0.3483\\
     CEM (ours) & +0.3589\\
     AECM (ours) & +0.3595\\
     CCM (ours) & \textbf{+0.3679}\\\hline
   \end{tabular}
   \caption{\label{rank_corr}Ablation and State of the art comparison with our models for Rank Correlation(higher is better)}
            
         

    \end{table}

\vspace{-0.5em}
\subsection{Dataset}
\vspace{-0.15cm}
We evaluate our proposed model on VQA Explanation Dataset (VQA-X)~\cite{huk_CVPR2018multimodal} which contains human annotated explanations for open-ended question answer(QA) pairs. QA pairs are taken from Visual Question Answering (VQA) dataset~\cite{VQA}. VQA-X consists of one explanation per QA pair in train split and three explanations per QA pair in validation and test split with a total of 31,536 explanations in training split, 4,377 explanations in validation split and 5904 explanations in test split. VQA-X also consists of human annotated visual justification collected from Amazon Mechanical Turk.
 \begin{table*}[htb]
   \centering

   \begin{tabular}{|l|c|c|c|c|c|c|c|c|}
   
     \hline
     \textbf{Model}&  \textbf{Combination}&\textbf{BLEU-4}& \textbf{METEOR}& \textbf{ROU-L}& \textbf{CIDERr}& \textbf{SPICE}\\\hline
        ME1~\cite{huk_CVPR2018multimodal}  &+ Ans + Att + Des  & 6.1 &12.8 &26.4 &36.2 &12.1\\
        ME2~\cite{huk_CVPR2018multimodal}  &- Ans - Att + Des  & 5.9 & 12.6 &26.3 &35.2 &11.9\\
        ME3 ~\cite{huk_CVPR2018multimodal}  &- Ans - Att + Exp  & 18.0 &17.3 &42.1 &63.6 &13.8\\
        ME4~\cite{huk_CVPR2018multimodal}  &+ Ans - Att + Exp  & 18.0 &17.6 &42.4 &66.3 &14.3\\
        ME5 ~\cite{huk_CVPR2018multimodal} &- Ans + Att + Exp  & 19.5 & 18.2 & 43.4 & 71.3 & 15.1\\
        ME6 ~\cite{huk_CVPR2018multimodal}  &+ Ans + Att + Exp  & 19.8 & {18.6} & 44.0 & {73.4} & {15.4}\\
        \hline
        Baseline &- Ans + Att + Exp&19.1 & 18.0 & 42.9 & 66.1 & 14.0\\
        Our (CCM) & + Ans + Att + Exp & \textbf{21.1} & \textbf{19.7} & \textbf{44.9} & \textbf{73.9} & \textbf{16.2}\\\hline 
   \end{tabular}
\vspace{-0.5em}
   \caption{\label{SOTA_reults} Comparison with State of the Art models on VQA-X dataset. Ans denotes use of ground truth answers for explanation generation, Att denotes answer attention and Exp denotes explanation attention. After using all three, our model outperforms all the other models.}
  \vspace{-1.2em} 
 \end{table*}


\begin{figure*}[htb]
	\centering
	\includegraphics[width=0.95\textwidth]{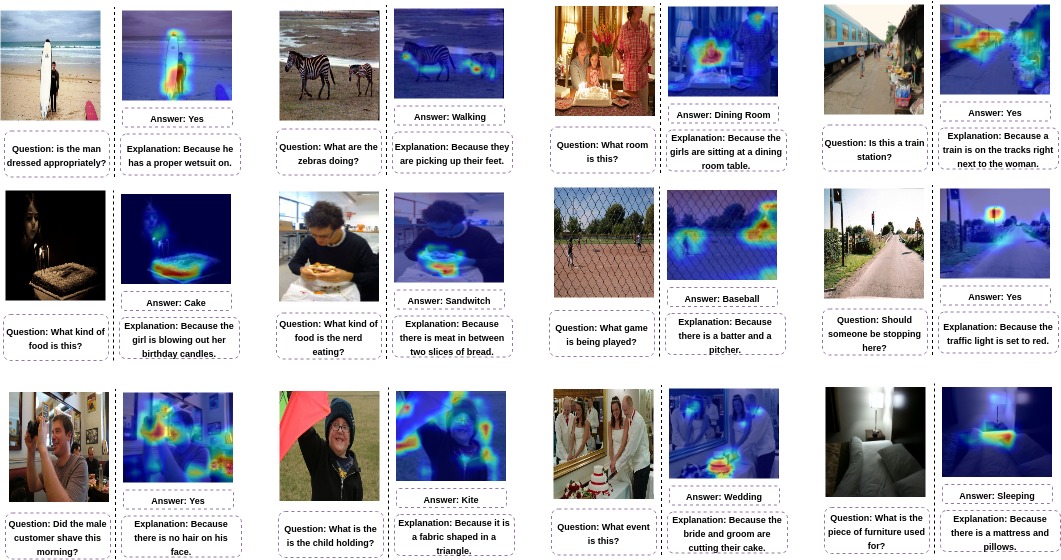}
  \vspace{-0.75em}
	\caption{This figure shows visual and textual explanation for images picked from validation set. The $1^{st}, 3^{rd}, 5^{th}$ and $7^{th}$ columns contain original image and its question. $2^{nd}, 4^{th}, 6^{th}$ and $8^{th}$ column provide predicted answer, visual explanation mask and textual explanation.}
	\label{fig:qualitative_results}
\end{figure*}
\vspace{-0.25cm}
\subsection{Ablation Analysis}
\vspace{-0.25cm}
We provide comparisons of our models with the prevalent baselines. We compare our model on textual and visual explanations.  We add noise during inference time to test the robustness of models. As the methods were not intended or designed to be robust to noise, normal methods perturbed more as compared to our models.

\subsubsection{Analysis on Textual Explanations}
 For textual explanations, we consider different variations of our method and various  ways to obtain collaborating embedding as mentioned in section ~\ref{mod:def}. Table ~\ref{abalation} shows the performance of variants of our model on different metrics for the VQA-X  test  set. It is clear that CCM outperforms all the other variants. There is a trend of increasing scores as we move from baseline to CCM. We achieve an improvement of about 2\% in BLEU-4, 1.7\% in METEOR, 2\% in ROUGE-L, 7\% in CIDEr and 2.1\% in SPICE.

 \begin{figure*}[ht]
     \small
     \centering
     \begin{tabular}[b]{ c  c  c}
     (a) Image  & (b) Question & (c) Image-Question \\ 
     \vspace{-0.1em}
     \includegraphics[width=0.32\textwidth]{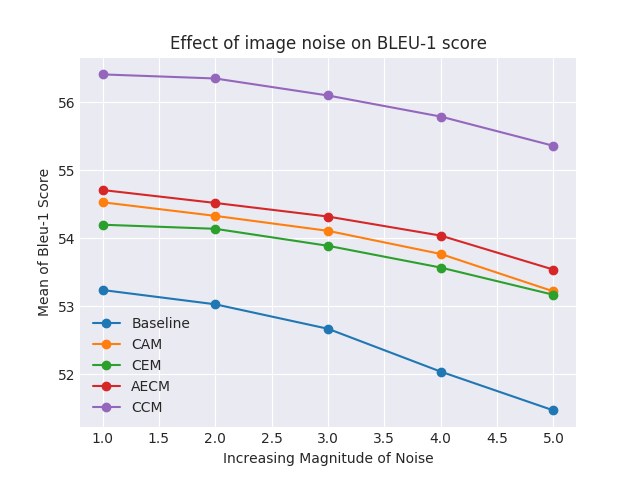}
     & \includegraphics[width=0.32\textwidth]{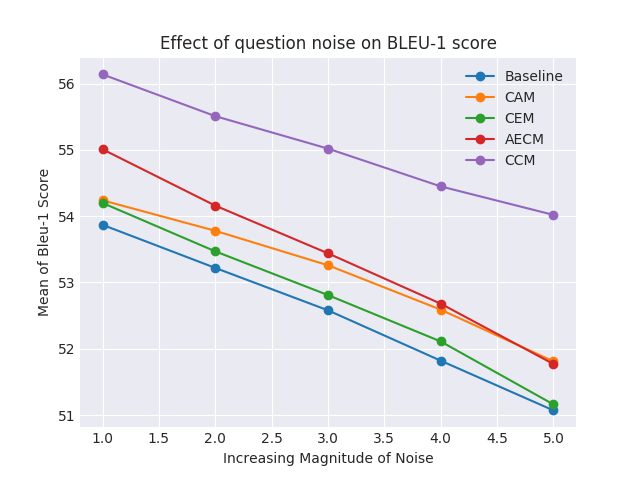}
     & \includegraphics[width=0.32\textwidth]{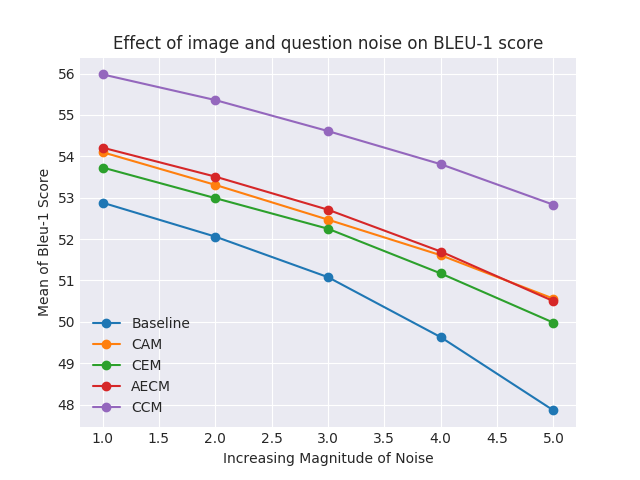}
       \end{tabular}
     \begin{tabular}[b]{ c  c  c}
     (a) Image  & (b) Question & (c) Image-Question \\ 
     
     \includegraphics[width=0.32\textwidth]{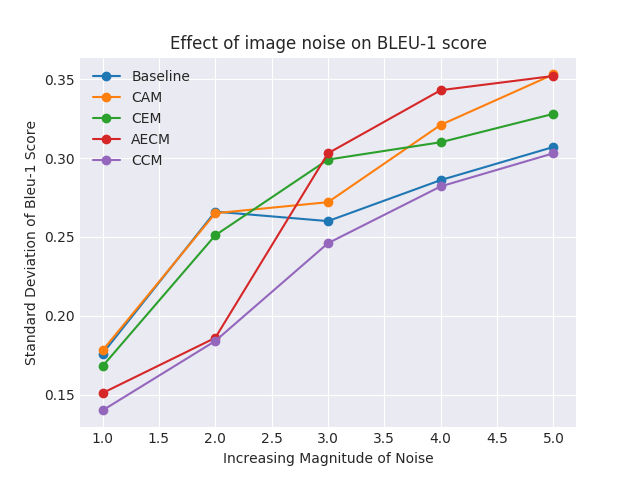}
     & \includegraphics[width=0.32\textwidth]{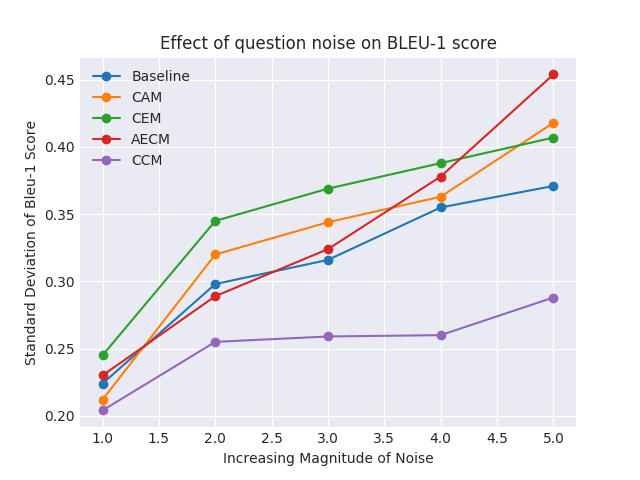}
     & \includegraphics[width=0.32\textwidth]{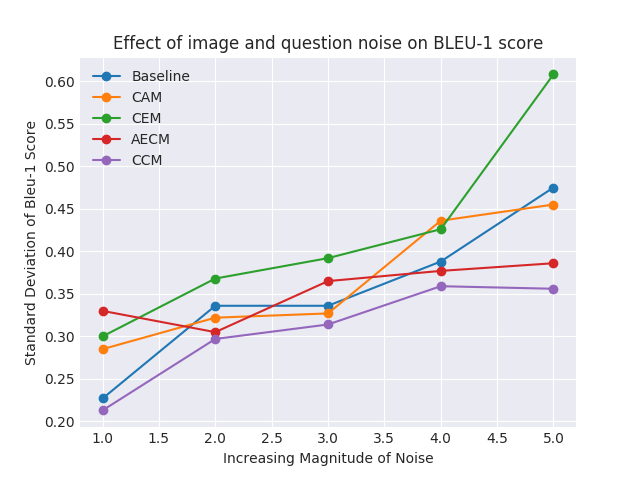}
       \end{tabular}
       \vspace{-0.4cm}
         \caption{\textbf{Effect of Image and Question Noise }: First row, First Column shows variation of Mean of Bleu-1 score for image noise, second column shows mean of Bleu-1 score for question noise and third column shows mean of Bleu-1 score for image and question noise combined. We add noise in image features by introducing gaussian noise with increasing noise intensity. Noise in questions is added by randomly masking question words with increasing probability of masking shown in column-2. Finally we add noise in both image features and question with increasing intensity/probability shown in column-3. Purple line shows performance of our model(CCM). It can be observed that our model perturbs least as compared to other models  and hence it is robust as compared to other models. Similarly second row provides the variation of Standard Deviation of Bleu-1 score for image noise, question noise and combined noise(image and question).}
          \label{tbl:mean}
     \vspace{-0.4cm}
 \end{figure*}
 
\subsubsection{Analysis on Visual Explanation }
     \vspace{-0.2cm}
We measure the similarity between our explanation and ground truth explanation by rank correlation and its results are shown in table~\ref{rank_corr}. We start with the rank correlation of the baseline model and then we compare with different variants such as CAM, CEM, AECM and CCM. We achieve an improvement of about 2.56\% when we move from state of the art(ME) to CCM. 

 \begin{figure}[ht]
	\centering
	\includegraphics[width=0.43\textwidth]{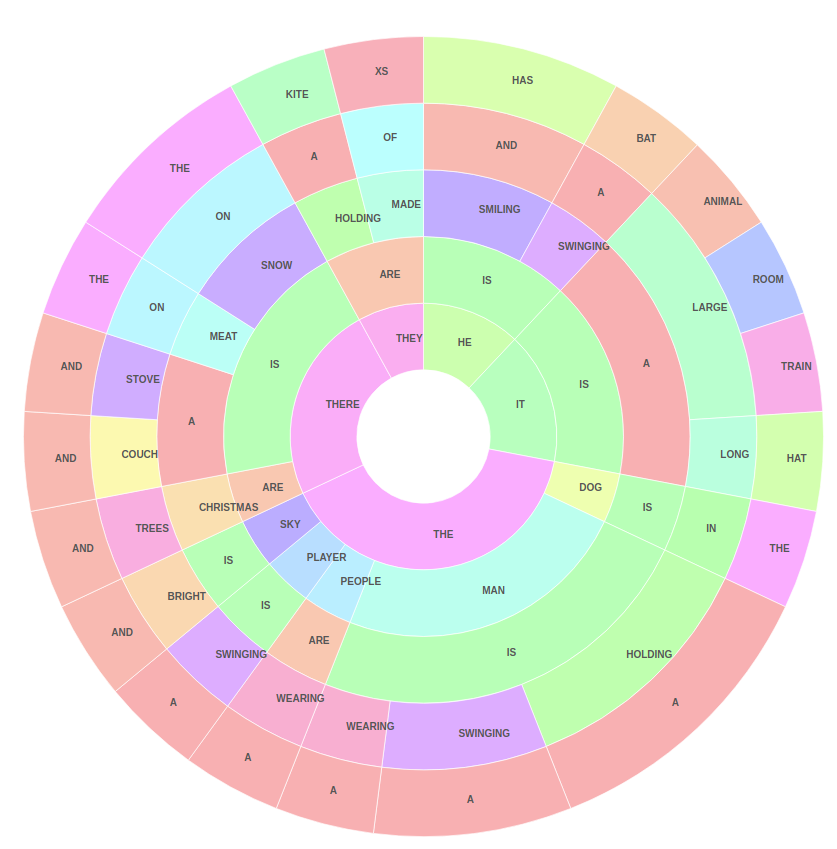}
	\vspace{-1em}
	\caption{Sunburst plot for Generated Explanation: The $i^{th}$ ring captures the frequency distribution over words for the $i^{th}$ word of the generated explanation. The angle subtended at the center is proportional to the frequency of the word. While some words have high frequency, the outer rings illustrate a fine blend of words. We have restricted the plot to 5 rings for easy readability.}
	\label{fig:sunburst}
	\vspace{-1.3em}
\end{figure}

\subsubsection{Statistical Significance Analysis}
     \vspace{-0.2cm}

\begin{figure}[ht]
	\centering
	\includegraphics[width=0.5\textwidth]{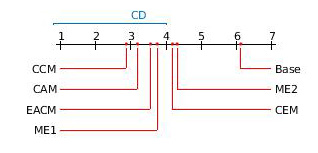}
	\vspace{-2em}
	\caption{ The mean rank of all the models on the basis of scores. Here CD=2.531 and p=0.007255. CDM refers to our model and others are different variations as described in section~\ref{abalation}. The colored lines between the two models represents that these models are not significantly different from each other.}
	\label{fig:SSA}
	   \vspace{-1.5em}
\end{figure}

We analyze Statistical Significance~\cite{Demvsar_JMLR2006} of our model(CCM) against the variants mentioned in section~\ref{SOTA_reults}. The Critical Difference(CD) for Nemenyi~\cite{Fivser_PLOS2016} test depends on given confidence level $\alpha$(0.05 in our case) for average ranks and number of tested datasets N. 
Low difference in ranks between two models implies that they are significantly less different and vice versa. Figure~\ref{fig:SSA} visualizes the post hoc analysis using the CD diagram. It is clear that CCM is significantly different from other methods. 

\begin{table}[htb]
\begin{center}
    \begin{tabular}{|l| c | c | c | c |} \hline
        \textbf{Model} & \textbf{Mean} & \textbf{Std Dev} & \textbf{Actual} & \textbf{Abs. Dif}\\\hline
        \textbf{Baseline} & 51.9 & 0.235 & 53.60 & 1.720\\
        \textbf{CAM}  & 53.9 & 0.254 & 54.42 & 0.520\\
        \textbf{CEM}  & 53.7 & 0.360 & 54.36 & 0.660\\
        \textbf{AECM}  & 54.3 & 0.257 & 54.39 & {0.093}\\
        \textbf{CCM} & 56.7 & \textbf{0.187} & 56.70 & \textbf{0.007}\\\hline
    \end{tabular}
\end{center}
  \vspace{-1.5em}
\caption{\label{image_question_noise_trained_with_noise}Performance of models against image and question noise(on BLUE-1 score)}
  \vspace{-1em}
\end{table}

 \vspace{-0.15cm}
\subsection{Analysis on Robustness of Model} 
\vspace{-0.15cm}
We analyze the behavior of our model against increasing noise in figure \ref{tbl:mean}. We sample the noise 50 times and report mean and standard deviation. Note that the model has not been trained with this noise, rather we only use it in the test time. Since Bleu-1 score varies the most(among all the scores), we choose Bleu-1 so that we can get proper estimates of standard deviation(Bleu-4 score deviates much less). First, we add gaussian noise to image features. Mean of the noise is same as that of the image features and the standard deviation is $\alpha \times (standard\ deviation)$ of the image features where $\alpha \in 1,..,5.$ We observe that our model has the highest mean, lowest standard deviation and lowest slope for mean and standard deviation. Second, we randomly mask words in questions during validation time. We mask words with increasing probabilities of $0.05, 0.1, 0.15, 0.2, 0.25$. Humans are robust to such noise since we can extract the semantic meaning of a sentence even if some words are masked or missing. So we test our model on this noise and observe that our model follows the previous trend. Then, we add both image and question noise with increasing magnitude. We again observe that our model outperforms all other models and it perturbs least.\\ 
In figure \ref{fig:replacing_words_image}(a), we analyze the effect of blurring the input images and extracting features from these noisy images. We observe that our model again outperforms other models along with deviating less w.r.t. the input. In figure \ref{fig:replacing_words_image}(b), we analyze the effect of replacing words in question from question vocabulary. Humans are not much affected by these since we can extract the meaning of a sentence based on the neighboring words. Again, our model outperforms all other models and perturbs the least. Since our model is learning joint distribution from two inputs(answers and explanations), it gives robust explanations for VQA.

 \vspace{-0.15cm}
\subsection{Comparison with State of the Art}
 \vspace{-0.15cm}
We obtain the initial comparison with the baselines on the rank correlation on visual explanation (VQA-X) dataset \cite{huk_CVPR2018multimodal} that provides a visual explanation in terms of segmentation mask while solving for VQA. We use a variant of the multimodal explanation \cite{huk_CVPR2018multimodal} model as our baseline method. We obtain an improvement of around 1.72\% using AECM network and 2.56\% using CCM network from state of the art model ME~\cite{huk_CVPR2018multimodal} in terms of rank correlation.  The comparison between various state-of-the-art methods and baselines is provided in table~\ref{rank_corr}. We compare our model with the state of the art model in table~\ref{SOTA_reults}. Rows 1 to 6 denote variants of state of the art model. The last row, CCM, is our best proposed model. We observe that for the VQA-X dataset, we achieve an improvement of 2\% in BLEU-4 score and 1.7\% in METEOR metric scores over the baseline.  We improve over the previous state-of-the-art ~\cite{huk_CVPR2018multimodal} for VQA-X dataset by around 1.3\% in BLEU-4 score and 1.1\% in METEOR score. 

 \vspace{-0.15cm}
\subsection{Qualitative Result}
 \vspace{-0.15cm}
We provide predicted answers, generated textual explanations and visual explanations in the form of attention maps for CCM in figure \ref{fig:qualitative_results}. It is apparent that our model points to prominent regions for answering questions as well as justifying the answer.  For example, in the first row, CCM is able to focus on a specific portion of the image and provide textual explanation for the predicted answer. The same results are observed for most of the examples. Also, we have shown the distribution of generated explanations, as demonstrated in the figure-\ref{fig:sunburst}.

 \vspace{-0.15cm}
\subsection{Training  and Model Configuration}
 \vspace{-0.15cm}
We use cross entropy loss to train Answer and Explanation Module and adversarial loss to train Correlated Module in an end-to-end manner. We use ADAM to update Answer and Explanation Module and SGD to update Correlated Module. We configure hyper-parameters of Generator module using the validation set which are as follows: {learning rate = 0.0007, batch size = 64, alpha = 0.99, beta = 0.95, and epsilon = 1e-8}. For correlated module, the hyper-parameters are as follows: {learning rate = 0.0007 and momentum = 0.9}. We also use weight clipping and learning rate decaying to decrease the learning rate every ten epochs.
 \vspace{-0.15cm}
\section{Conclusion}
 \vspace{-0.15cm}
In this paper, we solve for a method that can generate textual and visual explanations for visual question answering. The proposed method introduces a collaborative correlated module that ensures that generated textual explanations and answers are coherent with each other and are robust to perturbation in image and text. We provide a detailed evaluation that compares the accuracy of the answers and various standard metrics for the generated textual and visual explanations. We observe that while our model generates comparable accuracies for answer generation, the generated textual and visual explanations are more coherent and are robust against perturbation.  Our explanations change when there is a change in answer and hence signifies the importance of image and question in robustness of our model. 
In the future, we would consider ways for obtaining improved insights while solving challenging vision and language based tasks building upon the proposed work.

\section{Acknowledgements} Assistance through SERB grant CRG/2018/003566 is duly acknowledged.

{\small
\bibliographystyle{ieee}
\bibliography{egbib}

\begin{thebibliography}{10}\itemsep=-1pt

\bibitem{VQA}
S.~Antol, A.~Agrawal, J.~Lu, M.~Mitchell, D.~Batra, C.~L. Zitnick, and
  D.~Parikh.
\newblock {VQA}: {V}isual {Q}uestion {A}nswering.
\newblock In {\em International Conference on Computer Vision (ICCV)}, 2015.

\bibitem{Aytar_TPAMI2017}
Y.~Aytar, L.~Castrejon, C.~Vondrick, H.~Pirsiavash, and A.~Torralba.
\newblock Cross-modal scene networks.
\newblock {\em IEEE transactions on pattern analysis and machine intelligence},
  2017.

\bibitem{Banerjee_ACL2005}
S.~Banerjee and A.~Lavie.
\newblock Meteor: An automatic metric for mt evaluation with improved
  correlation with human judgments.
\newblock In {\em Proc. of ACL workshop on Intrinsic and Extrinsic Evaluation
  measures for Machine Translation and/or Summarization}, volume~29, pages
  65--72, 2005.

\bibitem{Barnard_JMLR2003}
K.~Barnard, P.~Duygulu, and D.~Forsyth.
\newblock N. de freitas, d.
\newblock {\em Blei, and MI Jordan," Matching Words and Pictures", submitted to
  JMLR}, 2003.

\bibitem{berg_ICCV2013you}
T.~Berg and P.~N. Belhumeur.
\newblock How do you tell a blackbird from a crow?
\newblock In {\em Proceedings of the IEEE International Conference on Computer
  Vision}, pages 9--16, 2013.

\bibitem{Chen_CVPR2015}
X.~Chen and C.~Lawrence~Zitnick.
\newblock Mind's eye: A recurrent visual representation for image caption
  generation.
\newblock In {\em Proceedings of the IEEE conference on computer vision and
  pattern recognition}, pages 2422--2431, 2015.

\bibitem{Das_ICCV2017}
A.~Das, S.~Kottur, J.~M. Moura, S.~Lee, and D.~Batra.
\newblock Learning cooperative visual dialog agents with deep reinforcement
  learning.
\newblock In {\em IEEE International Conference on Computer Vision (ICCV)},
  2017.

\bibitem{Demvsar_JMLR2006}
J.~Dem{\v{s}}ar.
\newblock Statistical comparisons of classifiers over multiple data sets.
\newblock {\em Journal of Machine learning research}, 7(Jan):1--30, 2006.

\bibitem{doersch_ACM2012makes}
C.~Doersch, S.~Singh, A.~Gupta, J.~Sivic, and A.~Efros.
\newblock What makes paris look like paris?
\newblock {\em ACM Transactions on Graphics}, 31(4), 2012.

\bibitem{escorcia_CVPR2015relationship}
V.~Escorcia, J.~Carlos~Niebles, and B.~Ghanem.
\newblock On the relationship between visual attributes and convolutional
  networks.
\newblock In {\em Proceedings of the IEEE Conference on Computer Vision and
  Pattern Recognition}, pages 1256--1264, 2015.

\bibitem{Fang_CVPR2015}
H.~Fang, S.~Gupta, F.~Iandola, R.~Srivastava, L.~Deng, P.~Doll{\'a}r, J.~Gao,
  X.~He, M.~Mitchell, J.~Platt, et~al.
\newblock From captions to visual concepts and back.
\newblock In {\em Proceedings of the IEEE conference on computer vision and
  pattern recognition}, 2015.

\bibitem{Farhadi_ECCV2010}
A.~Farhadi, M.~Hejrati, M.~A. Sadeghi, P.~Young, C.~Rashtchian, J.~Hockenmaier,
  and D.~Forsyth.
\newblock Every picture tells a story: Generating sentences from images.
\newblock In {\em European conference on computer vision}, pages 15--29.
  Springer, 2010.

\bibitem{Fivser_PLOS2016}
D.~Fi{\v{s}}er, T.~Erjavec, and N.~Ljube{\v{s}}i{\'c}.
\newblock Janes v0. 4: Korpus slovenskih spletnih uporabni{\v{s}}kih vsebin.
\newblock {\em Sloven{\v{s}}{\v{c}}ina}, 2(4):2, 2016.

\bibitem{Fukui_arXiv2016}
A.~Fukui, D.~H. Park, D.~Yang, A.~Rohrbach, T.~Darrell, and M.~Rohrbach.
\newblock Multimodal compact bilinear pooling for visual question answering and
  visual grounding.
\newblock {\em arXiv preprint arXiv:1606.01847}, 2016.

\bibitem{Gao_NIPS2015}
H.~Gao, J.~Mao, J.~Zhou, Z.~Huang, L.~Wang, and W.~Xu.
\newblock Are you talking to a machine? dataset and methods for multilingual
  image question.
\newblock In {\em Advances in Neural Information Processing Systems}, pages
  2296--2304, 2015.

\bibitem{goodfellow_NIPS2014}
I.~Goodfellow, J.~Pouget-Abadie, M.~Mirza, B.~Xu, D.~Warde-Farley, S.~Ozair,
  A.~Courville, and Y.~Bengio.
\newblock Generative adversarial nets.
\newblock In {\em Advances in neural information processing systems}, pages
  2672--2680, 2014.

\bibitem{Goyal_CVPR2017}
Y.~Goyal, T.~Khot, D.~Summers-Stay, D.~Batra, and D.~Parikh.
\newblock Making the v in vqa matter: Elevating the role of image understanding
  in visual question answering.
\newblock In {\em Proceedings of the IEEE Conference on Computer Vision and
  Pattern Recognition}, pages 1--9, 2017.

\bibitem{Guo_arxiv2017long}
J.~Guo, S.~Lu, H.~Cai, W.~Zhang, Y.~Yu, and J.~Wang.
\newblock Long text generation via adversarial training with leaked
  information.
\newblock {\em arXiv preprint arXiv:1709.08624}, 2017.

\bibitem{Hu_ICML2017toward}
Z.~Hu, Z.~Yang, X.~Liang, R.~Salakhutdinov, and E.~P. Xing.
\newblock Toward controlled generation of text.
\newblock In {\em International Conference on Machine Learning}, pages
  1587--1596, 2017.

\bibitem{huk_CVPR2018multimodal}
D.~Huk~Park, L.~Anne~Hendricks, Z.~Akata, A.~Rohrbach, B.~Schiele, T.~Darrell,
  and M.~Rohrbach.
\newblock Multimodal explanations: Justifying decisions and pointing to the
  evidence.
\newblock In {\em Proceedings of the IEEE Conference on Computer Vision and
  Pattern Recognition}, pages 8779--8788, 2018.

\bibitem{jain_CVPR2017}
U.~Jain, Z.~Zhang, and A.~Schwing.
\newblock Creativity: Generating diverse questions using variational
  autoencoders.
\newblock {\em arXiv preprint arXiv:1704.03493}, 2017.

\bibitem{Johnson_CVPR2016}
J.~Johnson, A.~Karpathy, and L.~Fei-Fei.
\newblock Densecap: Fully convolutional localization networks for dense
  captioning.
\newblock In {\em Proceedings of the IEEE Conference on Computer Vision and
  Pattern Recognition}, pages 4565--4574, 2016.

\bibitem{Karpathy_CVPR2015}
A.~Karpathy and L.~Fei-Fei.
\newblock Deep visual-semantic alignments for generating image descriptions.
\newblock In {\em Proceedings of the IEEE conference on computer vision and
  pattern recognition}, pages 3128--3137, 2015.

\bibitem{Kulkarni_CVPR2011}
G.~Kulkarni, V.~Premraj, S.~Dhar, S.~Li, Y.~Choi, A.~C. Berg, and T.~L. Berg.
\newblock Baby talk: Understanding and generating image descriptions.
\newblock In {\em Proceedings of the 24th CVPR}. Citeseer, 2011.

\bibitem{lane_USC2005explainable}
H.~C. Lane, M.~G. Core, M.~Van~Lent, S.~Solomon, and D.~Gomboc.
\newblock Explainable artificial intelligence for training and tutoring.
\newblock Technical report, UNIVERSITY OF SOUTHERN CALIFORNIA MARINA DEL REY CA
  INST FOR CREATIVE~…, 2005.

\bibitem{li_arxiv2017adversarial}
J.~Li, W.~Monroe, T.~Shi, A.~Ritter, and D.~Jurafsky.
\newblock Adversarial learning for neural dialogue generation.
\newblock {\em arXiv preprint arXiv:1701.06547}, 2017.

\bibitem{Li_NIPS2016}
R.~Li and J.~Jia.
\newblock Visual question answering with question representation update (qru).
\newblock In {\em Advances in Neural Information Processing Systems}, pages
  4655--4663, 2016.

\bibitem{liang_CORR2017recurrent}
X.~Liang, Z.~Hu, H.~Zhang, C.~Gan, and E.~P. Xing.
\newblock Recurrent topic-transition gan for visual paragraph generation.
\newblock {\em CoRR, abs/1703.07022}, 2, 2017.

\bibitem{Lin_ACL2004}
C.-Y. Lin.
\newblock Rouge: A package for automatic evaluation of summaries.
\newblock In {\em Text summarization branches out:Proceedings of the ACL-04
  workshop}, 2004.

\bibitem{Lin_ECCV2014}
T.-Y. Lin, M.~Maire, S.~Belongie, J.~Hays, P.~Perona, D.~Ramanan,
  P.~Doll{\'a}r, and C.~L. Zitnick.
\newblock Microsoft coco: Common objects in context.
\newblock In {\em European Conference on Computer Vision}, pages 740--755.
  Springer, 2014.

\bibitem{Lu_NIPS2016}
J.~Lu, J.~Yang, D.~Batra, and D.~Parikh.
\newblock Hierarchical question-image co-attention for visual question
  answering.
\newblock In {\em Advances In Neural Information Processing Systems}, pages
  289--297, 2016.

\bibitem{Ma_AAAI2016}
L.~Ma, Z.~Lu, and H.~Li.
\newblock Learning to answer questions from image using convolutional neural
  network.
\newblock In {\em Thirtieth AAAI Conference on Artificial Intelligence}, 2016.

\bibitem{Malinowski_NIPS2014}
M.~Malinowski and M.~Fritz.
\newblock A multi-world approach to question answering about real-world scenes
  based on uncertain input.
\newblock In {\em Advances in Neural Information Processing Systems (NIPS)},
  2014.

\bibitem{mirza_arxiv2014conditional}
M.~Mirza and S.~Osindero.
\newblock Conditional generative adversarial nets.
\newblock {\em arXiv preprint arXiv:1411.1784}, 2014.

\bibitem{Mostafazadeh_ACL2016}
N.~Mostafazadeh, I.~Misra, J.~Devlin, M.~Mitchell, X.~He, and L.~Vanderwende.
\newblock Generating natural questions about an image.
\newblock {\em arXiv preprint arXiv:1603.06059}, 2016.

\bibitem{Noh_CVPR2016}
H.~Noh, P.~Hongsuck~Seo, and B.~Han.
\newblock Image question answering using convolutional neural network with
  dynamic parameter prediction.
\newblock In {\em Proceedings of the IEEE Conference on Computer Vision and
  Pattern Recognition}, pages 30--38, 2016.

\bibitem{Papineni_ACL2002}
K.~Papineni, S.~Roukos, T.~Ward, and W.-J. Zhu.
\newblock Bleu: a method for automatic evaluation of machine translation.
\newblock In {\em Proceedings of the 40th annual meeting on association for
  computational linguistics}, pages 311--318. Association for Computational
  Linguistics, 2002.

\bibitem{Patro_2018_CVPR}
B.~Patro and V.~P. Namboodiri.
\newblock Differential attention for visual question answering.
\newblock In {\em The IEEE Conference on Computer Vision and Pattern
  Recognition (CVPR)}, June 2018.

\bibitem{Patro2019ExplanationVA}
B.~N. Patro, Anupriy, and V.~P. Namboodiri.
\newblock Explanation vs attention: A two-player game to obtain attention for
  vqa.
\newblock {\em ArXiv}, abs/1911.08618, 2019.

\bibitem{Patro_EMNLP2018MDN}
B.~N. Patro, S.~Kumar, V.~K. Kurmi, and V.~Namboodiri.
\newblock Multimodal differential network for visual question generation.
\newblock In {\em Proceedings of the 2018 Conference on Empirical Methods in
  Natural Language Processing}, pages 4002--4012. Association for Computational
  Linguistics, 2018.

\bibitem{Patro2019_ICCV2019UCAM}
B.~N. Patro, M.~Lunayach, S.~Patel, and V.~P. Namboodiri.
\newblock U-cam: Visual explanation using uncertainty based class activation
  maps.
\newblock {\em ICCV}, abs/1908.06306, 2019.

\bibitem{radford2015unsupervised}
A.~Radford, L.~Metz, and S.~Chintala.
\newblock Unsupervised representation learning with deep convolutional
  generative adversarial networks.
\newblock {\em arXiv preprint arXiv:1511.06434}, 2015.

\bibitem{Reed_CVPR2016}
S.~Reed, Z.~Akata, X.~Yan, L.~Logeswaran, B.~Schiele, and H.~Lee.
\newblock Generative adversarial text to image synthesis.
\newblock {\em arXiv preprint arXiv:1605.05396}, 2016.

\bibitem{Ren_NIPS2015}
M.~Ren, R.~Kiros, and R.~Zemel.
\newblock Exploring models and data for image question answering.
\newblock In {\em Advances in Neural Information Processing Systems (NIPS)},
  pages 2953--2961, 2015.

\bibitem{Shih_CVPR2016}
K.~J. Shih, S.~Singh, and D.~Hoiem.
\newblock Where to look: Focus regions for visual question answering.
\newblock In {\em Proceedings of the IEEE Conference on Computer Vision and
  Pattern Recognition}, pages 4613--4621, 2016.

\bibitem{shortliffe_MB1975model}
E.~H. Shortliffe and B.~G. Buchanan.
\newblock A model of inexact reasoning in medicine.
\newblock {\em Mathematical biosciences}, 23(3-4):351--379, 1975.

\bibitem{Socher_TACL2014}
R.~Socher, A.~Karpathy, Q.~V. Le, C.~D. Manning, and A.~Y. Ng.
\newblock Grounded compositional semantics for finding and describing images
  with sentences.
\newblock {\em Transactions of the Association of Computational Linguistics},
  2(1):207--218, 2014.

\bibitem{Vedantam_CVPR2015}
R.~Vedantam, L.~Zitnick, and D.~Parikh.
\newblock Cider: Consensus-based image description evaluation.
\newblock In {\em IEEE Conference on Computer Vision and Pattern Recognition
  (CVPR)}, pages 4566--4575, 2015.

\bibitem{Velivckovic_SSCI2016}
P.~Veli{\v{c}}kovi{\'c}, D.~Wang, N.~D. Lane, and P.~Li{\`o}.
\newblock X-cnn: Cross-modal convolutional neural networks for sparse datasets.
\newblock In {\em Computational Intelligence (SSCI), 2016 IEEE Symposium Series
  on}, pages 1--8. IEEE, 2016.

\bibitem{Vijayakumar_2016diverse}
A.~K. Vijayakumar, M.~Cogswell, R.~R. Selvaraju, Q.~Sun, S.~Lee, D.~Crandall,
  and D.~Batra.
\newblock Diverse beam search: Decoding diverse solutions from neural sequence
  models.
\newblock {\em arXiv preprint arXiv:1610.02424}, 2016.

\bibitem{Vinyals_CVPR2015}
O.~Vinyals, A.~Toshev, S.~Bengio, and D.~Erhan.
\newblock Show and tell: A neural image caption generator.
\newblock In {\em Proceedings of the IEEE Conference on Computer Vision and
  Pattern Recognition}, pages 3156--3164, 2015.

\bibitem{Xu_ECCV2016}
H.~Xu and K.~Saenko.
\newblock Ask, attend and answer: Exploring question-guided spatial attention
  for visual question answering.
\newblock In {\em European Conference on Computer Vision}, pages 451--466.
  Springer, 2016.

\bibitem{Xu_ICML2015}
K.~Xu, J.~Ba, R.~Kiros, K.~Cho, A.~Courville, R.~Salakhudinov, R.~Zemel, and
  Y.~Bengio.
\newblock Show, attend and tell: Neural image caption generation with visual
  attention.
\newblock In {\em International Conference on Machine Learning}, pages
  2048--2057, 2015.

\bibitem{Yan_ECCV2016}
X.~Yan, J.~Yang, K.~Sohn, and H.~Lee.
\newblock Attribute2image: Conditional image generation from visual attributes.
\newblock In {\em European Conference on Computer Vision}, pages 776--791.
  Springer, 2016.

\bibitem{Yu_ICCV2015}
L.~Yu, E.~Park, A.~C. Berg, and T.~L. Berg.
\newblock Visual madlibs: Fill in the blank description generation and question
  answering.
\newblock In {\em Computer Vision (ICCV), 2015 IEEE International Conference
  on}, pages 2461--2469. IEEE, 2015.

\bibitem{Yu_AAAI2017seqgan}
L.~Yu, W.~Zhang, J.~Wang, and Y.~Yu.
\newblock Seqgan: Sequence generative adversarial nets with policy gradient.
\newblock In {\em AAAI}, pages 2852--2858, 2017.

\bibitem{zeiler_ECCV2014visualizing}
M.~D. Zeiler and R.~Fergus.
\newblock Visualizing and understanding convolutional networks.
\newblock In {\em European conference on computer vision}, pages 818--833.
  Springer, 2014.

\bibitem{zhang_arxiv2017adversarial}
Y.~Zhang, Z.~Gan, K.~Fan, Z.~Chen, R.~Henao, D.~Shen, and L.~Carin.
\newblock Adversarial feature matching for text generation.
\newblock {\em arXiv preprint arXiv:1706.03850}, 2017.

\bibitem{zhou_arxiv2014object}
B.~Zhou, A.~Khosla, A.~Lapedriza, A.~Oliva, and A.~Torralba.
\newblock Object detectors emerge in deep scene cnns.
\newblock {\em arXiv preprint arXiv:1412.6856}, 2014.

\bibitem{Zhu_CVPR2016}
Y.~Zhu, O.~Groth, M.~Bernstein, and L.~Fei-Fei.
\newblock Visual7w: Grounded question answering in images.
\newblock In {\em Proceedings of the IEEE Conference on Computer Vision and
  Pattern Recognition}, pages 4995--5004, 2016.

\end{thebibliography}
}

\end{document}